\documentclass[letterpaper]{article} % DO NOT CHANGE THIS
\usepackage{aaai2026}  % DO NOT CHANGE THIS
\usepackage{times}  % DO NOT CHANGE THIS
\usepackage{helvet}  % DO NOT CHANGE THIS
\usepackage{courier}  % DO NOT CHANGE THIS
\usepackage[hyphens]{url}  % DO NOT CHANGE THIS
\usepackage{graphicx} % DO NOT CHANGE THIS
\usepackage{natbib}  % DO NOT CHANGE THIS AND DO NOT ADD ANY OPTIONS TO IT
\usepackage{caption} % DO NOT CHANGE THIS AND DO NOT ADD ANY OPTIONS TO IT
\usepackage{algorithm}
\usepackage{algorithmic}
\usepackage{amsmath}
\usepackage{amsfonts}
\usepackage{array}
\usepackage{booktabs}
\usepackage{multirow}
\usepackage{newfloat}
\usepackage{listings}
\DeclareCaptionStyle{ruled}{labelfont=normalfont,labelsep=colon,strut=off} % DO NOT CHANGE THIS
\floatstyle{ruled}
\newfloat{listing}{tb}{lst}{}
\floatname{listing}{Listing}
\title{ESMC: MLLM-Based Embedding Selection for Explainable Multiple Clustering}
\author{
    Xinyue Wang\textsuperscript{\rm 1}, 
    Yuheng Jia\textsuperscript{\rm 1,\rm 2, \rm3}\thanks{Yuheng Jia is the corresponding author.},
    Hui Liu\textsuperscript{\rm 3}, 
    Junhui Hou\textsuperscript{\rm 4}
}
\affiliations{
    \textsuperscript{\rm 1}School of Computer Science and Engineering, Southeast University, Nanjing 210096, China \\
    \textsuperscript{\rm 2}Key Laboratory of New Generation Artificial Intelligence Technology and Its Interdisciplinary Applications  \\ (Southeast University), Ministry of Education, China \\
    \textsuperscript{\rm 3}Department of Computing and Information Sciences, Saint Francis University, Hong Kong, China \\
    \textsuperscript{\rm 4}Department of Computer Science, City University of Hong Kong, Hong Kong, China \\

    213222801@seu.edu.cn, yhjia@seu.edu.cn,
    h2liu@sfu.edu.hk, jh.hou@cityu.edu.hk
}

\usepackage{bibentry}
\begin{document}

\maketitle

\begin{abstract}
Typical deep clustering methods, while achieving notable progress, can only provide one clustering result per dataset. This limitation arises from their assumption of a fixed underlying data distribution, which may fail to meet user needs and provide unsatisfactory clustering outcomes. Our work investigates how multi-modal large language models (MLLMs) can be leveraged to achieve user-driven clustering, emphasizing their adaptability to user-specified semantic requirements. However, directly using MLLM output for clustering has risks for producing unstructured and generic image descriptions instead of feature-specific and concrete ones. To address these issues, our method first discovers that MLLMs' hidden states of text tokens are strongly related to the corresponding features, and leverages these embeddings to perform clusterings from any user-defined criteria. We also employ a lightweight clustering head augmented with pseudo-label learning, significantly enhancing clustering accuracy. Extensive experiments demonstrate its competitive performance on diverse datasets and metrics. Codes and datasets are available in github: 
\end{abstract}

\begin{links}
    \link{Code}{https://github.com/JCSTARS/Embedding-Selective-Multiple-Clustering}
\end{links}

\section{Introduction}
\begin{figure*}[t]
    \centering
\includegraphics[width=0.95\linewidth]{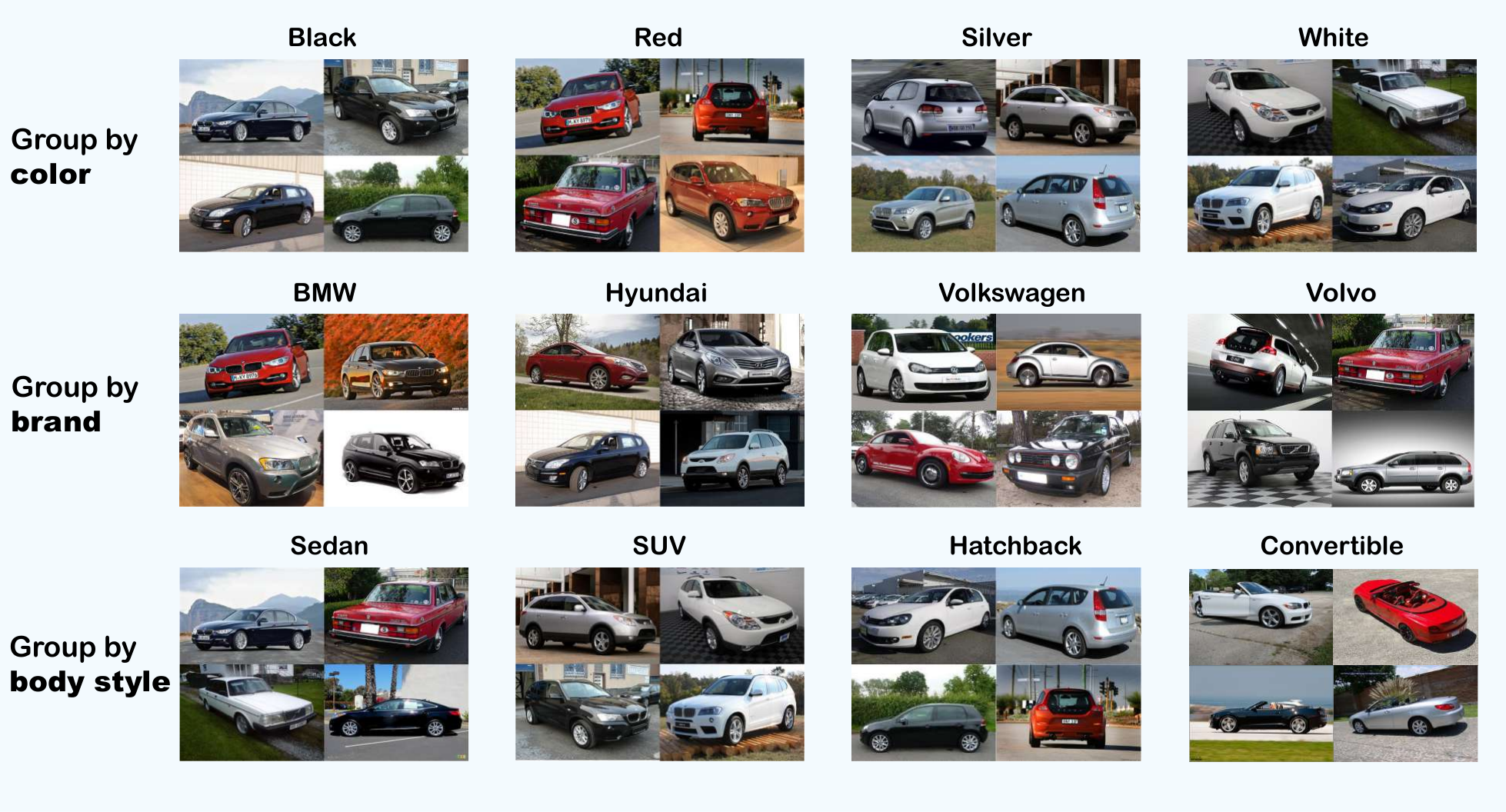}
    \caption{Multiple clustering of sample images on the Stanford\_cars dataset with cluster criterion of color, manufacturer, and body style.}
    \label{multi_dis}
\end{figure*}

Clustering is an unsupervised machine learning technique focused on grouping objects based on their specific patterns \citep{jain1999data,xu2005survey}. Traditional clustering methods \citep{macqueen1967some, ester1996density} often rely on hand-crafted features, which can face difficulties with real-world, high-dimensional data. Deep clustering \citep{van2020scan,ren2024deep,caron2018deep} addresses this by leveraging the strong representation learning power of deep neural networks to produce feature embeddings that are more conducive to clustering. Most of these methods group the dataset with a single standard. However, users may require clustering a set of images based on multiple criteria, as illustrated in Figure \ref{multi_dis}. For instance, a set of car images can be categorized according to different attributes such as color, brand, or body style. This scenario highlights the challenge of multi-clustering, where diverse grouping objectives coexist for the same dataset.

Multiple clustering demonstrates that a single dataset can be partitioned into distinct clusters based on different underlying structures \citep{yu2024multiple}. While existing methods have offered solutions for obtaining diverse clusters \citep{bae2006coala,dasgupta2010mining,ren2022diversified,yao2023augdmc}, they often rely on inherent data characteristics and lack the flexibility to directly incorporate user-defined criteria.

Foundational models like CLIP \citep{radford2021learning} recently enable clustering tasks \citep{yao2024multi,yao2024customized} by aligning image and text features. However, CLIP's prompt-agnostic image embeddings limit clustering based on semantic features. On the other hand, MLLMs like LLaVA \citep{liu2023visual,liu2024improved} leverage LLMs and visual instruction tuning, enabling user-defined clustering through modality alignment.

\begin{figure*}[t]
    \centering
\includegraphics[width=\linewidth]{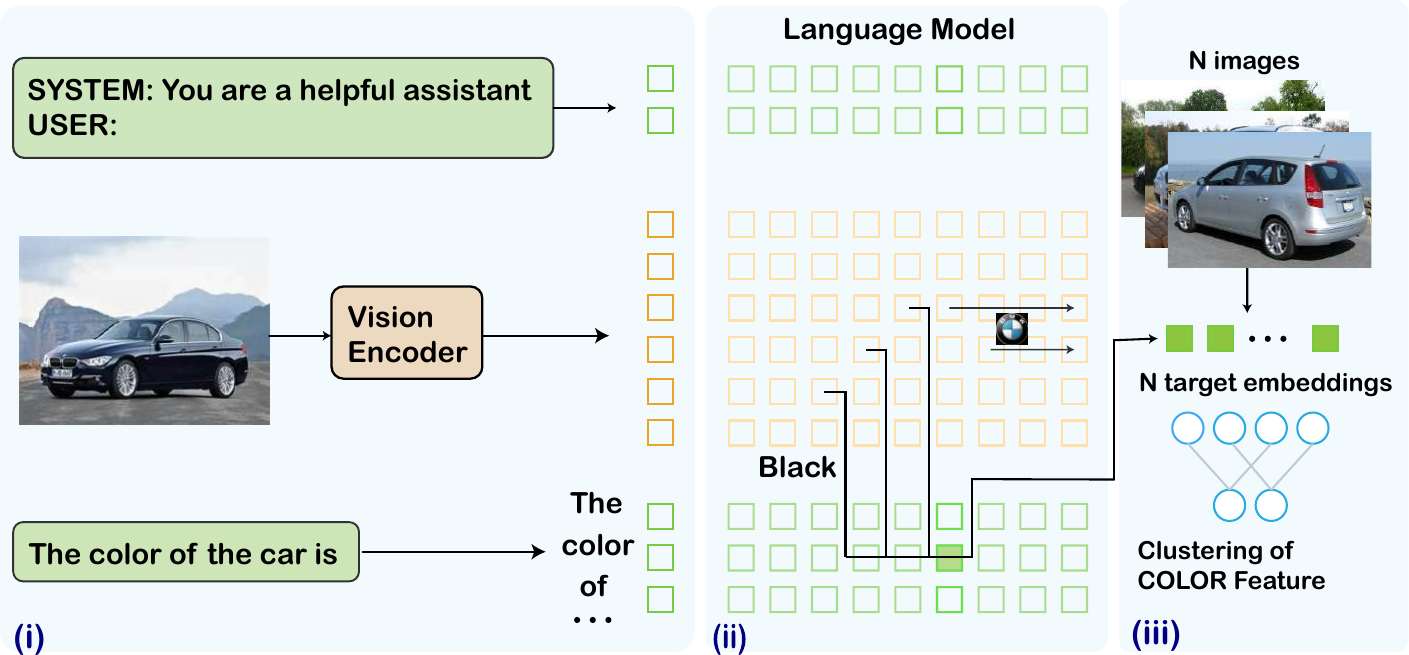}
    \caption{The ESMC framework consists of three key components: (i) the MLLM processes multimodal inputs (images and text prompts); (ii) target embeddings are selected from the language model's hidden states corresponding to text tokens; and (iii) a lightweight clustering head is trained to enhance cluster accuracy.}
    \label{pipeline}
\end{figure*}

However, naively using MLLMs for multiple clustering can face several challenges. Firstly, MLLMs tend to produce unstructured data \citep{li2024large}, making it difficult to link semantically similar outputs to specific clustering labels. Furthermore, the model's response may offer a too general image description instead of strictly following provided instructions. A comprehensive discussion of these problems is presented in Section~\ref{subsec:4.3}. To mitigate these issues, we propose Embedding Selective Multiple Clustering (ESMC). As shown in Figure \ref {pipeline}, ESMC first derives internal embeddings of text prompt tokens to extract features. These embeddings are from hidden states of text tokens of the language model in MLLMs. As visualized in Figure~\ref{logit}, color-related features are strongly related to the `color' prompt token, while prompt-unrelated or image-unrelated tokens present low logits. We also employ a clustering head with pseudo-label learning to group these embeddings into well-separated clusters.

Our contributions can be summarized as follows.
\begin{itemize}
    \item We reveal that the internal embeddings of text prompt tokens can serve as condensed representations of corresponding features and can be utilized as an effective basis for multiple clustering.
    \item We demonstrate that a lightweight 2-layer MLP clustering head with pseudo label supervision achieves substantial improvements.
    \item Extensive experiments and ablation studies indicate that the proposed approach provides user-defined, diverse, and high-quality clustering outcomes.
\end{itemize}

\section{Related Work}

\paragraph{Multiple clustering} Multiple clustering aims to discover various meaningful ways to group a dataset \citep{yu2024multiple}. Traditional methods often achieve this by partitioning data based on its internal characteristics, such as constructing orthogonal subspaces \citep{cui2007non,niu2013iterative} or modeling independent mixture distributions \citep{jain2008simultaneous,tokuda2017multiple}, employing objective functions to improve both cluster quality and diversity \citep{yu2024multiple}. Recent deep learning advancements have also offered improved solutions using techniques like autoencoders, muMeasuring statistical dependence with Hilbert-Schmidt normsitep{yao2023augdmc} to learn better representations for diverse clustering. For example, MCV \citep{guerin2018improving} utilizes different pre-trained feature extractors as diverse views for the same data to construct multiple clusters. ENRC \citep{miklautz2020deep} finds multiple non-redundant clusters in the embedded space of the autoencoder. iMClusts \citep{ren2022diversified} leverages multi-head attention to generate significant subspaces and Hilbert Schmidt Independence Criterion (HSIC) \citep{gretton2005measuring} to minimize their dependencies. Multi-Map and Multi-Sub utilize CLIP \citep{radford2021learning} embeddings and proxy learning to generate multiple clusters. Most previous methods cannot provide clusters based on arbitrary user-defined criteria.

\paragraph{Multi-modal large language models} MLLMs typically integrate three core components: a vision encoder for extracting visual features, a projection network to align visual and text modalities, and an autoregressive language backbone. Taking LLaVA-1.5 \citep{liu2024improved} as an example, its architecture comprises CLIP \citep{radford2021learning} as the vision encoder, a projection module, and Vicuna-v1.5 \citep{zheng2023judging} as the language foundation. Given an input image $I$, the CLIP image encoder $f_{vision\_encoder}$ will provide visual vectors. A mapping network $f_{proj}$ is applied to project image features and text features to the same space and provides $X_{visual}$ tokens: 
\begin{equation}
  \mathbf{X}_{visual}=f_{proj}(f_{vision\_enc}(I))\in \mathbb{R}^{n\times d_{mllm}},
\end{equation}
where $d_{mllm}$ is the dimension of MLLM embedding.
The input prompt $S$ is mapped into $m$ tokens through the tokenizer $f_{tokenizer}$ to word embeddings from $f_{embed}$:
\begin{equation}
\textbf{X}_{text} = f_{embed}(f_{tokenizer}(S)) \in \mathbb{R}^{m \times d_{mllm}}.
\end{equation}
The concatenated modality embeddings $\left[\mathbf{X}_{vis}; \mathbf{X}_{text}\right]$ are fed into the transformer-based language model. Let $A_l(k) \in \mathbb{R}^{d_{model}}$ denote the $k$-th token's representation at the $l$-th transformer layer, and $A_0=\left[\mathbf{X}_{vis}; \mathbf{X}_{text}\right]$. The final output is generated from $A_{L} \in \mathbb{R}^{{(m+n+t)}\times d_{model}}$ through an unembedding matrix $\mathbf{W}u \in \mathbb{R}^{d{model} \times |V|}$, where $|V|$ is the vocabulary size to map the embeddings to the vocabulary space, and $t$ represents special tokens.

\paragraph{Cracking the Internal Mechanisms of MLLMs} Detecting and understanding the internal workings of Large Language Models (LLMs) has long been recognized as crucial for trustworthy AI \citep{dai2021knowledge,zou2024improving,zhao2024explainability}. Inspired by these works, initial efforts have explored the interpretability of Multi-modal Large Language Models (MLLMs). For instance, Vl-interp \citep{jiang2024interpreting} employs logit lens to identify and edit object hallucinations. LLaVA-interp \citep{neo2024towards} demonstrates a strong spatial correlation between object information and its original image location, indicating a refinement of visual input into interpretable language tokens. LVLM-interp \citep{stan2024lvlm} introduces interactive tools such as attention maps and causal graphs to aid in LVLM interpretation. However, these investigations have primarily focused on the visual processing aspects of LVLMs, leaving the internal mechanisms related to prompt token embeddings unexplored, which motivates our work. 

The logit lens \citep{nostalgebraist2020logitlens, belrose2023eliciting, pal2023future}  serves as an interpretability technique for analyzing language models by projecting intermediate layer representations through the unembedding matrix $\mathbf{W}_u$. Specifically, it computes linguistic probabilities via:
\begin{equation}
E_l(k) = \mathbf{W}_u \cdot A_l(k) \in \mathbb{R}^{|V|},
\label{eq5}
\end{equation}
where $E_l(k)$ denotes the $k$-th embedding feature of $l$-th layer in vocabulary space. While logit lens has been utilized as a tool to understand LLMs' hidden states, our proposed method first extends its application to explain prompt tokens and their semantic relationships with visual features in MLLMs.

\section{Multiple Clustering with MLLM Hidden Embedding and Lightweight Clustering Head}
We leverage user-defined prompts to specify clustering criteria and present our proposed pipeline as a two-stage framework:

\paragraph{Target embedding extraction} Based on our novel observation of feature-related embeddings from text token embeddings, we sample a few images to get precise target embeddings, as described in detail in Section~\ref{subsec:3.2}.
% Identification target embeddings corresponding tÍo user-specified semantic attributes.
\paragraph{Clustering head training} A lightweight clustering head is trained to enhance clustering performance. This network transforms raw embeddings into a structured latent space suitable for clustering, which is elaborated in Section~\ref{subsec:3.3}.

\subsection{Find Target Embeddings}
\label{subsec:3.2}
\begin{figure*}[t]
    \centering
\includegraphics[width=1.0\linewidth]{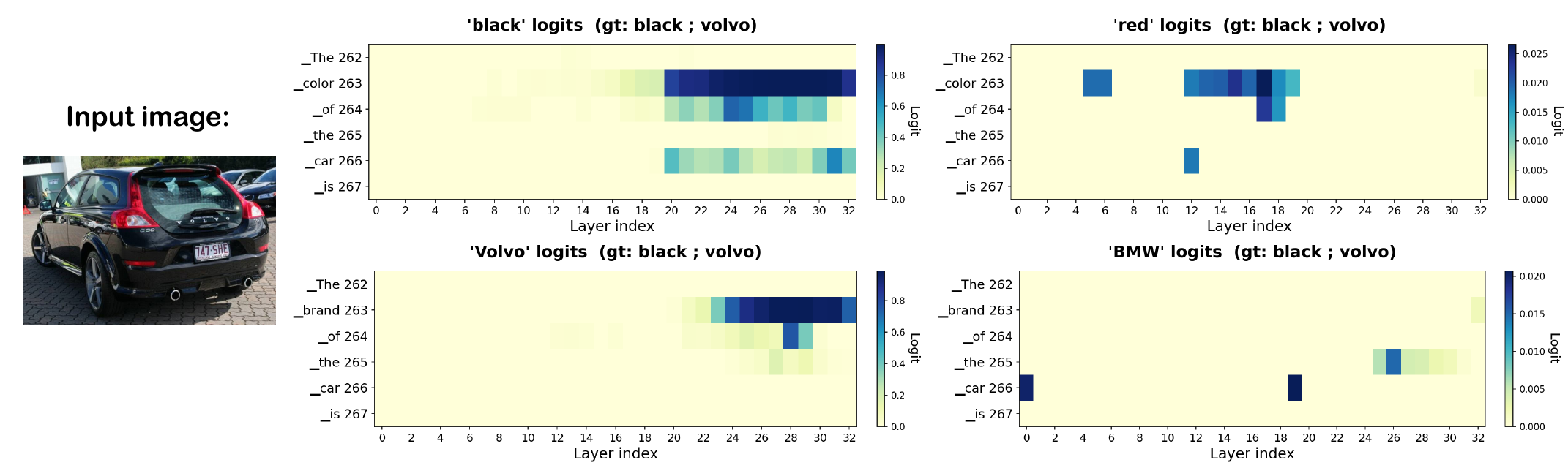}
    \caption{The relationship between vocabulary logits and text prompt tokens, note that the color bar varies across the figures. The image is from the Stanford\_cars dataset.}
    \label{logit}
\end{figure*}

We observe that specific prompt token shows strong correlations with semantically related visual features. The positions of these embeddings are determined by the input prompt tokens. For example, as shown in Figure~\ref{logit}, the ground-truth labels for color and brand features are ``black" and ``Volvo", respectively. The color-related feature of the input image (e.g., ``black") correlates strongly with the embeddings at position $E_{l}(263)$, which corresponds to the ``color" prompt token. Similarly, the brand-related feature (e.g., ``Volvo") also shows high logits in the embeddings at position $E_l(263)$, aligning with the ``brand" prompt token. Conversely, features not present in the image, such as ``red" for color or ``BMW" for brand, exhibit significantly lower logits in their respective embeddings, enabling clear distinctions between different features. 
Additional examples are provided in Appendix \ref{apd:more_examples}. Further analysis on the robustness of the prompt chosen can be found in Appendix \ref{subsec:prompt_robustness}.

Though selecting embeddings from later layers corresponding to specific input tokens provides reasonable performance, we propose a refined approach to enhance clustering accuracy and semantic feature representation. Specifically, we recommend sampling a small set of images, like 10 examples for each feature, to identify embeddings with high contextual specificity, followed by 2 steps:

\paragraph{Keyword generation} We use GPT-4 to generate feature-relevant keywords (e.g., asking ``What are common car colors?" and GPT would provide tokens like ``white", ``black", ``blue", ``red", etc, as a response). These keywords are crucial to locate the embedding positions.

\paragraph{Embedding localization} For each keyword, we track the embeddings of the sampled images with top-$k$ logit and choose the final target embeddings from those with top-$k$ logit shared by different keywords. For example, we choose the ``black" keyword generated by GPT-4, and black cars are included in the sample of images, as indicated in Figure~\ref{logit}. The ``black" token should present high logits in embeddings like $E_{27}(263), E_{28}(263), E_{29}(263)$ and $ E_{30}(263)$. These embeddings are candidates for target embeddings. We define logit values exceeding 0.2 as high logits~\cite{jiang2024interpreting}. More details regarding the impact of sampled image quantity and Keyword generation on performance can be found in Appendix \ref{subsec:sampling}.

\subsection{Pseudo Label Clustering Head training}
\label{subsec:3.3}
While typical clustering methods like K-means \citep{macqueen1967some} can be applied directly to target embeddings from user-defined prompt criteria, we observe that raw hidden state embeddings from LLMs/MLLMs often exhibit suboptimal performance in clustering tasks \citep{petukhova2025text}. To mitigate this limitation, our approach incorporates a two-layer MLP clustering head to align high-dimensional embeddings with low-level semantic clusters.

The procedure is formalized as follows:
\paragraph{Initialization} Clusters are initialized via K-means. Given $\mathcal{X} = {x_1, x_2, \dots, x_N}$ where $x_i \in \mathbb{R}^{|V|}$, we initialize clusters using K-means:
\begin{equation}
{C_1, C_2, \dots, C_K} = \arg\min_{{C_k}} \sum_{k=1}^K \sum_{x_i \in C_k} \|x_i - \mu_k\|^2,
\end{equation}
where $K$ is the given number of target clusters. $N$ is the number of clustering samples. $\mu_k \in \mathbb{R}^d$ is the centroid of cluster $C_k$. $x_i$ is the target embedding of $E_l(k)$

\paragraph{Pseudo-label generation} For each cluster $C_j$ with centroid $\mu_j$, calculate the squared Euclidean distance between each data point $x_i \in C_j$ and its cluster centroid:
\begin{equation}
d_{ij}=||x_i - \mu_j||^2.
\end{equation}
For each cluster $C_j$, select the top $\alpha$ percent of data points that have the smallest distances $d_{ij}$ to the cluster centroid $\mu_j$. Let $S_j$ be the set of these selected data points from cluster $C_j$, where $\alpha$ is a hyperparameter. Assign the pseudo-label $y_i=j$ to each selected data point $x_i \in S_j$. This creates a pseudo-labeled training set:
\begin{equation}
D_{pseudo}={(x_i,y_i)\mid x_i\in \bigcup_{j=1}^kSj}.
\end{equation}

\paragraph{Network training} The MLP is trained using these pseudo-labeled samples with cross-entropy loss to optimize clustering accuracy. The loss function can be denoted as:
\begin{equation}
L = -\frac{1}{k} \sum_{i=1}^{k} \sum_{j=1}^{C} y_{ij} \log(\hat{y}_{ij}).
\end{equation}

\section{Experiments}

\subsection{Experiment Setup}
\label{subsec:4.1}
\paragraph{Datasets}
We evaluate our method on seven multi-clustering benchmark datasets: Stanford\_Cars \citep{yao2024customized} includes two criteria: \textit{color} (red, black, white, silver) and \textit{type} (BMW, Hyundai, Volkswagen, Volvo). Flower dataset \citep{yao2024customized} contains: \textit{species} (daisy, hyacinth, lily, rose) and \textit{color} (pink, purple, red, white, yellow). Fruit \citep{hu2017finding} also contains two clustering criteria, \textit{color} (red, green, yellow) and \textit{species} (apple, banana, grape). Card \citep{yao2023augdmc} includes two criteria, \textit{number} (ace to king) and \textit{suits} (spades, clubs, hearts, diamonds). CMU\_face \citep{gunnemann2014smvc} comprises three criteria, \textit{emotion} (happy, sad, neutral, angry), \textit{pose} (left, right, up, straight), and \textit{glasses presence} (wearing sunglasses or not). Fruit360 \citep{yao2023augdmc} extends the Fruit dataset with additional samples: \textit{color} (red, green, yellow, burgundy) and \textit{species} (apple, banana, grape, cherry). CIFAR10  \citep{yao2024multi}features two criteria, \textit{environment} (sea, sky, land) and \textit{object type} (animals, transportation). For the Flower and Stanford\_Cars datasets, we extend prior work \citep{yao2024multi} by constructing multi-criteria benchmarks based on dataset descriptions. The datasets are openly released with our codes.

\paragraph{Implementation details} Target embeddings are extracted from the hidden states of LLaVA-1.5-7b \citep{liu2024improved} during inference, while user-defined criteria are encoded via prompts such as ``The manufacturer of the car is", ``The species of the fruit is", and ``The color of the flower is." The pseudo-label ratio hyperparameter $\alpha$ is empirically set to 0.1 for the CMU\_face and Card dataset, 0.2 for Fruit360 and CIFAR10 datasets, 0.3 for Stanford\_cars and Flower datasets, and  0.4 for Fruit datasets. The clustering head is implemented as a two-layer MLP, with the first layer mapping the input dimension $|V|$ to 512 units and the second layer projecting these features to the target cluster number. We train the clustering head for 100 epochs. We employ Normalized Mutual Information (NMI) \citep{meilua2007comparing} and the Rand Index (RI) \citep{rand1971objective} as external metrics to measure the similarity between clustering results and ground truth. The experiments are conducted on one RTX3090 card with 24GB of memory.

\subsection{Results}

\begin{table*}[htb!]
\centering
\setlength{\tabcolsep}{0.85mm} 
\begin{tabular}{
  l 
  l |
  *{8}{
>{\centering\arraybackslash}p{1.55cm} 
  }
}
\toprule
Dataset & Criteria & MCV & ENRC & iMClusts & AugDMC & DDMC & Multi-Map & Multi-Sub & \textbf{ESMC} \\
\midrule
\multirow{2}{*}{Stanford\_Cars} 
& Color & 0.2103 & 0.2465 & 0.2336 & 0.2736 & 0.6899 & 0.7360 & \underline{0.7533} & \textbf{0.8138} \\
& Type & 0.1650 & 0.2063 & 0.1963 & 0.2364 & 0.6045 & 0.6355 & \underline{0.6616} & \textbf{0.8817} \\
\midrule
\multirow{2}{*}{Flowers} 
& Color & 0.2938 & 0.3329 & 0.3169 & 0.3556 & 0.6327 & 0.6426 & \underline{0.6940} & \textbf{0.7283} \\
& Species & 0.1326 & 0.1561 & 0.1894 & 0.1887 & 0.1996 & 0.6148 & \underline{0.7580} & \textbf{0.8923} \\
\midrule
\multirow{2}{*}{CIFAR-10} 
& Type & 0.1618 & 0.1826 & 0.2040 & 0.2855 & 0.3991 & 0.4967 & \underline{0.5271} & \textbf{0.8293} \\
& Environment & 0.1379 & 0.1892 & 0.1920 & 0.2927 & 0.3782 & 0.4598 & \underline{0.4828} & \textbf{0.6075} \\
\midrule
\multirow{3}{*}{CMU\_face} 
& Emotion & 0.1433 & 0.1592 & 0.0422 & 0.0161 & 0.1726 & 0.1786 & \underline{0.2053} & \textbf{0.2237} \\
& Glass & 0.1201 & 0.1493 & 0.1299 & 0.1039 & 0.2261 & 0.3402 & \underline{0.4870} & \textbf{0.7665} \\
& Pose & 0.3254 & 0.2290 & 0.4437 & 0.1320 & 0.4526 & 0.4693 & \underline{0.5923} & \textbf{0.6271} \\
\midrule
\multirow{2}{*}{Card} 
& Order & 0.0792 & 0.1225 & 0.1144 & 0.1440 & 0.1563 & 0.3633 & \underline{0.3921} & \textbf{0.4736} \\
& Suits & 0.0430 & 0.0676 & 0.0716 & 0.0873 & 0.0933 & 0.2734 & \underline{0.3104} & \textbf{0.3586} \\
\midrule
\multirow{2}{*}{Fruit360} 
& Color & 0.3777 & 0.4264 & 0.4097 & 0.4594 & 0.4981 & 0.6239 & 0.6654 & \textbf{0.6952} \\
& Species & 0.2985 & 0.4142 & 0.3861 & 0.5139 & \underline{0.5292} & 0.5284 & \textbf{0.6123} & 0.5036 \\
\midrule
\multirow{2}{*}{Fruit} 
& Color & 0.6266 & 0.7103 & 0.7351 & 0.8517 & 0.8973 & 0.8619 & \textbf{0.9693} & \underline{0.9308} \\
& Species & 0.2733 & 0.3187 & 0.3029 & 0.3546 & 0.3764 & 1.0000 & \underline{1.0000} & \textbf{1.0000} \\
\bottomrule
\end{tabular}

\caption{Quantitative comparison of 8 approaches using NMI. For methods involving k-means, the average result of 10 times is reported.}

\label{nmi-res}
\end{table*}

\begin{table*}[htb!]
\centering

\setlength{\tabcolsep}{0.85mm} 
\begin{tabular}{
  l
  l|
  *{8}{
>{\centering\arraybackslash}p{1.55cm} 
  }
}
\toprule
Dataset & Criteria & MCV & ENRC & iMClusts & AugDMC & DDMC & Multi-Map & Multi-Sub & \textbf{ESMC} \\

\midrule
\multirow{2}{*}{Stanford\_Cars} & Color & 0.5802 & 0.6779 & 0.6552 & 0.7525 & 0.8765 & 0.9193 & \textbf{0.9387} & \underline{0.9235} \\
& Type & 0.5634 & 0.6217 & 0.5643 & 0.7356 & 0.7957 & 0.8399 & \underline{0.8792} & \textbf{0.9589} \\
\midrule
\multirow{2}{*}{Flowers} & Color & 0.5860 & 0.6214 & 0.6127 & 0.6931 & 0.7887 & 0.7984 & \underline{0.8843} & \textbf{0.8992} \\
& Species & 0.5273 & 0.6065 & 0.6195 & 0.6077 & 0.6227 & 0.8321 & \underline{0.8980} & \textbf{0.9726} \\
\midrule
\multirow{2}{*}{CIFAR-10} & Type & 0.5634 & 0.6217 & 0.5643 & 0.7356 & 0.7957 & 0.8399 & \underline{0.7394} & \textbf{0.9589} \\
& Environment & 0.3344 & 0.3599 & 0.3664 & 0.4689 & 0.5547 & 0.6737 & \underline{0.7096} & \textbf{0.8282} \\
\midrule
\multirow{4}{*}{CMU\_face} & Emotion & 0.5268 & 0.6630 & 0.5932 & 0.5367 & 0.7593 & \underline{0.7105} & \textbf{0.8527} & 0.6697 \\
& Glass & 0.4955 & 0.6209 & 0.5627 & 0.5361 & 0.7663 & 0.7068 & \underline{0.8324} & \textbf{0.9259} \\
& Pose & 0.6028 & 0.5029 & 0.6114 & 0.5517 & 0.7904 & 0.6624 & \textbf{0.8736} & \underline{0.8253} \\
\midrule
\multirow{2}{*}{Card} & Order & 0.7128 & 0.7313 & 0.7658 & 0.8267 & 0.8326 & 0.8587 & \textbf{0.8842} & \underline{0.8805} \\
& Suits & 0.3638 & 0.3801 & 0.3715 & 0.4228 & 0.6469 & 0.7039 & \textbf{0.8504} & \underline{0.7285} \\
\midrule
\multirow{2}{*}{Fruit360} & Color & 0.6791 & 0.6868 & 0.6841 & 0.7392 & 0.7472 & 0.8439 & \textbf{0.8821} & \underline{0.8493} \\
& Species & 0.6176 & 0.6984 & 0.6732 & 0.7430 & \underline{0.7703} & 0.7582 & \textbf{0.8504} & 0.7285 \\
\midrule
\multirow{2}{*}{Fruit} & Color & 0.7685 & 0.8511 & 0.8632 & 0.9108 & 0.9383 & 0.9556 & \textbf{0.9964} & \underline{0.9732} \\
& Species & 0.6597 & 0.6536 & 0.6743 & 0.7399 & 0.7621 & 1.0000 & \underline{1.0000} & \textbf{1.0000} \\
\bottomrule
\end{tabular}

\caption{Quantitative comparison of 8 approaches using RI. For methods involving k-means, the average result of 10 times is reported.}
\label{main-res-ri}
\end{table*}

As presented in Tables~\ref{nmi-res} and~\ref{main-res-ri}, the proposed ESMC method demonstrates overall superior performance compared to baselines across various clustering tasks and datasets, evidenced by its NMI and RI scores. On CIFAR10-Type, ESMC achieved the top NMI (0.8293) and RI (0.9520), marking a significant improvement over Multi-Sub (0.5271 for NMI, 0.7394 for RI) and Multi-Map (0.4967 for NMI, 0.7104 for RI). ESMC also exceeded baseline results for Fruit-Species, CMU\_face-Glass, Flower dataset, CIFAR-10 dataset, and Stanford\_Cars-Type datasets. It obtained perfect scores for Fruit-Species and achieved leading scores of NMI 0.8923 and RI 0.9726 on Flower-Species, surpassing Multi-Sub's NMI by 21.09\%.

We note minor underperformance in the Fruit360-Color NMI and Fruit NMI scores. This may arise from the characteristics of the datasets, such as unclear images, and differences in the internal knowledge of the underlying models (e.g., LLaVA and CLIP). For instance, our analysis, detailed in Appendix \ref{sec:appendix_failure_cases}, reveals instances where models misidentified colors or object types due to ambiguous visual cues. 
Previous methods like Multi-Map and Multi-Sub, often relying on specific keywords, exhibit fewer such ambiguities.

\begin{figure}[h]
    \centering
\includegraphics[width=\linewidth]{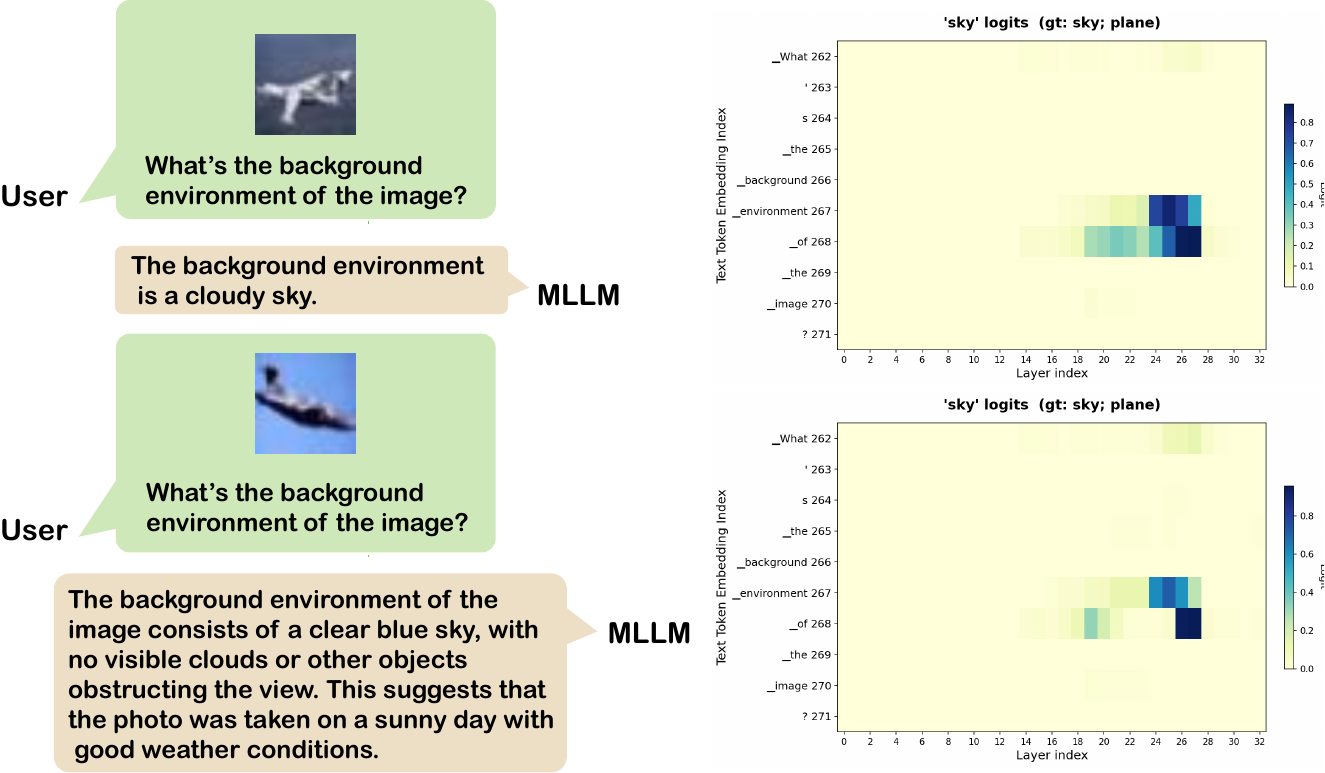}
    \caption{Left: Two MLLM outputs with shared semantic meaning but distinct formatting. Right: The target embedding we choose $E_{25}(267)$, shows higher consistency in ``sky" logit, as the clustering label in the environment criterion.}
    \label{unstruct}
\end{figure}

\begin{figure}[h]
    \centering
\includegraphics[width=\linewidth]{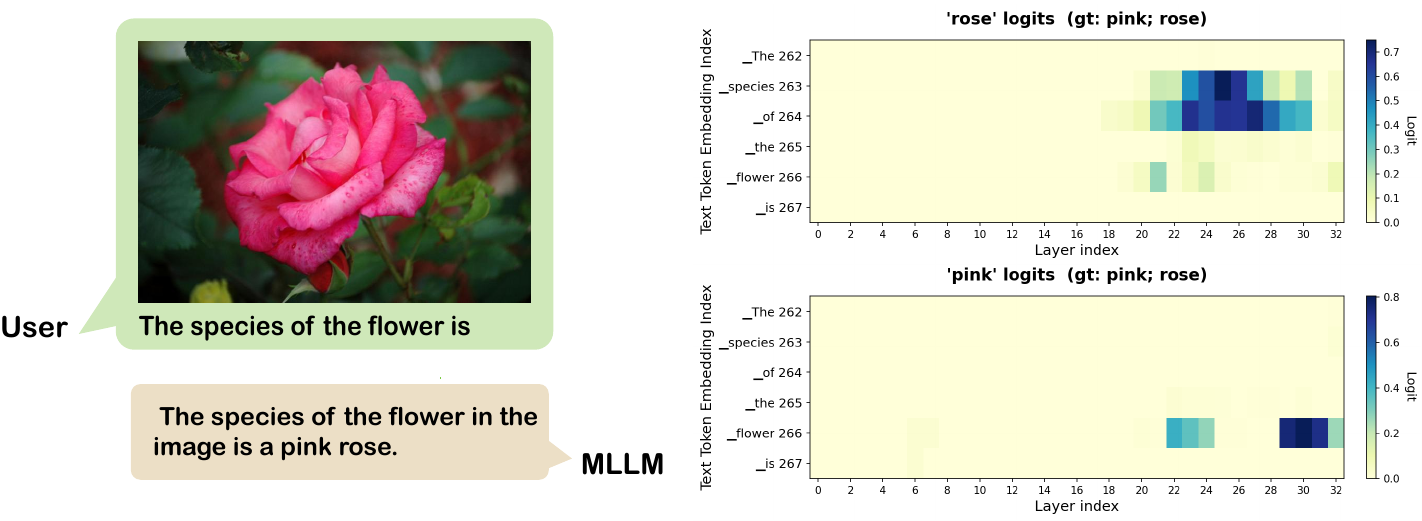}
    \caption{Left:  Conversation reveals that the MLLM response can be too general. It responds as ``pink rose" instead of ``rose" directly, while we only care about the species feature instead of the color feature. Right: Our target embedding $E_{25}(263)$ shows high logit in ``rose" logit and low in ``pink" logit, which could indicate the species feature.}
    \label{general}
\end{figure}

\subsection{Ablation Studies}
\label{subsec:4.3}
\paragraph{Clustering of MLLM outputs yields suboptimal performance}
To evaluate the limitations of clustering directly by the MLLM linguistic output, we compare our method with a baseline approach that clusters text outputs from LLaVA. Specifically, we first extract textual outputs from the model and generate semantic embeddings by feeding them into the CLIP text encoder \citep{radford2021learning}. Experimental results (Table~\ref{com-cap}) confirm that our proposed approach outperforms this clustering strategy. The inferior performance of the baseline arises from three key factors. Firstly, the unstructured nature of MLLM outputs introduces inconsistent responses to inputs. As shown in Figure~\ref{unstruct}, when presented with two images of the same environment type described as ``The background environment is a cloudy sky" and a much longer description, the CLIP text encoder may fail to produce semantically aligned embeddings. In contrast, MLLM embeddings from ESMC assign nearly identical logits to the ``sky" token (0.8194 and 0.7436), which suggests consistency in semantic representation, particularly after applying the softmax function to embedding logits. These two images are highly likely to be clustered together due to their shared environmental features.

\begin{table*}[htb!]
    \centering
    \begin{tabular}{@{}ll|ccc|ccc@{}}
        \toprule \centering
        Dataset & Criteria & \multicolumn{3}{c}{NMI↑} & \multicolumn{3}{c}{RI↑} \\ \cmidrule(lr){3-5} \cmidrule(lr){6-8} 
        & & Output & ESMC, w/o.MLP & \textbf{ESMC} & Output & ESMC, w/o.MLP & 
        \textbf{ESMC} \\ \midrule 
        \multirow{2}{*}{Stanford\_Cars} & Color & 0.7871 & \underline{0.7934} & \textbf{0.8138} & 0.9107 & \underline{0.9129}&\textbf{0.9235}\\
        & Type & 0.6462 &\underline{0.8694} & \textbf{0.8817} & 0.8474 & \underline{0.9517} &\textbf{0.9589} \\
        \midrule
        \multirow{2}{*}{Flower}  & Color & 0.6698 &\underline{0.6835}& \textbf{0.7283} & 0.8736 & \underline{0.8810} & \textbf{0.8982} \\
        & Species & 0.8133 & \underline{0.8462} &\textbf{0.8923} & 0.9329 & \underline{0.9347} &\textbf{0.9726} \\ 
        \midrule
        \multirow{2}{*}{CIFAR-10} 
        & Type & \underline{0.7505} & 0.7149 & \textbf{0.8293} & 0.8953& \underline{0.9137} & \textbf{0.9520} \\
        & Environment & 0.0443  & \underline{0.5484} & \textbf{0.6075} & 0.4941 & \underline{0.7860} &\textbf{0.8282}\\
        \midrule
        \multirow{3}{*}{CMU\_face} & Emotion & 0.1498 & \underline{0.1902} & \textbf{0.2237} & 0.5876 &  \underline{0.6346} &\textbf{0.6697} \\
        & Glass & \underline{0.6780} & 0.6445 & \textbf{0.7665} & \underline{0.8279} & 0.7955 &\textbf{0.9259}\\
        & Pose & 0.0913 & \underline{0.5498} & \textbf{0.6271} & 0.5631 & \underline{0.8140}  &\textbf{0.8253}\\
        \midrule
        \multirow{2}{*}{Card} & Order & \underline{0.4387} &0.3895& \textbf{0.4736} & 0.8348 & \underline{0.8712} & \textbf{0.8805} \\
        & Suits & 0.2423 & \underline{0.3334} & \textbf{0.3586} & 0.4258 & \underline{0.6982} &\textbf{0.7258}\\
        \midrule
        \multirow{2}{*}{Fruit360} 
        & Color & 0.6071 & \underline{0.6523} & \textbf{0.6952} & \underline{0.7972} & 0.7584 & \textbf{0.8493}\\
        & Species & 0.2199 & \underline{0.3806} &\textbf{0.5036}  & 0.5080 & \underline{0.6824} &\textbf{0.7285}\\
        \midrule
        \multirow{2}{*}{Fruit} & Color& 0.9308  & 0.9308 & 0.9308 & 0.9732 & 0.9732 & 0.9732\\
        & Species & 0.8441 & \underline{1.0000} & \textbf{1.0000}  & 0.9337 & \underline{1.0000} & \textbf{1.0000} \\
        \bottomrule
    \end{tabular}
        \caption{The table compares three clustering approaches: (1) Output (direct clustering using the model's linguistic output), (2) Ours, w/o.MLP (clustering without an MLP-based clustering head), and (3) ESMC (The proposed approach). We do not present the results of adding the clustering head after the MLLM output with CLIP embeddings due to significant performance degradation. }
    \label{com-cap}
\end{table*}

\begin{table*}[htb]
    \centering
\setlength{\tabcolsep}{1.2mm} 
    \begin{tabular}{c|ccccccc}
    \toprule \centering
  Dataset & Stanford\_cars & Flower & CIFAR-10 & CMU\_face & Card & Fruit360 & Fruit \\  \midrule
time/s & 0.3675 & 0.4472 & 15.03 & 0.5016 & 4.636 & 2.874 & 0.3968 \\
\bottomrule
\end{tabular} 
    \caption{Total training time (100 epochs with different features) for clustering heads.}
    \label{time}
\end{table*}

Secondly, MLLM tends to provide overly general descriptions instead of specific features. As illustrated in Figure~\ref{general}, when prompted with ``The species of the flower is", the model responds ``pink rose" instead of focusing on the species feature. This conflates distinct attributes like color and species, making it harder to cluster based on precise criteria.

Thirdly, language bias \citep{he2024cracking,leng2024mitigating,lee2023volcano}—MLLM's tendency to prioritize prior linguistic knowledge over visual context—directly contributes to performance gaps. This bias creates a critical trade-off: excessive reliance on language priors may induce hallucinations, which happen when MLLMs generate outputs inconsistent with images, while underweighting linguistic guidance risks losing discriminative information encoded in the input prompts.

The proposed approach, ESMC, leverages intermediate hidden states from earlier layers of the MLLM, rather than the final output layer. This design not only preserves the text prompt's semantic guidance through embeddings at special positions encoded by quantized probability but also mitigates language bias by avoiding the dominance of higher-level of linguistic abstractions in later auto-regressive layers, which are prone to over-reliance on previously generated responses. By doing so, we ensure that our target embeddings are grounded in the initial, raw fusion of visual and linguistic information, while with less linguistic bias or language priors\citep{lin2023revisiting} from the luanguage model itself. 

\paragraph{Clustering head improves performance}
Clustering head \citep{van2020scan, niu2022spice,jia2024towards} is a lightweight and effective technique for enhancing clustering performance. Our results, presented as NMI and RI gains in Table~\ref{com-cap}, show that the inclusion of a clustering head can lead to substantial improvements, reaching up to 12\% on the Fruit360 dataset of species feature and the CMU\_face dataset of Glass feature. This benefit is likely due to the high dimensionality of the MLLM embeddings after projection into the vocabulary space, causing the curse of dimensionality\citep{indyk1998approximate,donoho2000high}.  Although ESMC embeddings provide richer feature representations, they seem less directly suited for clustering. Notably, we observed that the clustering head did not enhance performance when applying CLIP embeddings, likely because these embeddings already exhibit a strong inherent capacity for clustering in semantic spaces. The clustering head is also highly time-efficient, with the only trainable part is the 2-layer MLP head, resulting in significantly smaller FLOPs and memory usage compared to previous method\cite{yao2024multi}, which is detailed in Table~\ref{time}.

\section{Conclusion and Limitation}
In conclusion, our proposed method, ESMC, introduces an explainable framework for multiple clustering tasks, addressing limitations of existing multimodal language models (MLLMs). While MLLMs excel in visual feature extraction and contextual knowledge, they can yield unstructured or overly generic interpretations. To mitigate this, ESMC obtains target embeddings via sampling and employs a lightweight clustering head with pseudo-label learning for well-separated clusters. The method's transparency is grounded in the use of specific, interpretable text token embeddings as the clustering basis. By focusing the clustering process on these features, ESMC not only leverages the strengths of MLLMs but also enhances the interpretability of their internal representations. We also present the limitations and future directions.

\paragraph{Specific language model architecture} 
Based on our study, MLLMs like LLaVA, built on LLaMA (Touvron et al. 2023), exhibit specific text prompt embedding properties. We found this architecture type fuses text and image features at a relatively late stage, with vision features injected as visual tokens.

\paragraph{Future work} 
While our work focused on applying text prompt embeddings to multiple clustering tasks, we believe MLLMs offer other effective ways for clustering or retrieval. Our research may also aid MLLM interpretability and grounding techniques. Ultimately, we aim for this work to foster more reliable and safer MLLM.

\bibliography{aaai2026}

\newpage
\appendix
\section{Appendices}
We include the source code for our experiments, a hyperparameter sensitivity analysis, prompt robustness analysis, and additional examples of target embedding selection and illustrations.

\subsection{Hyperparameter analysis}
\begin{figure}[htb]
	\centering
	\includegraphics[width=\linewidth]{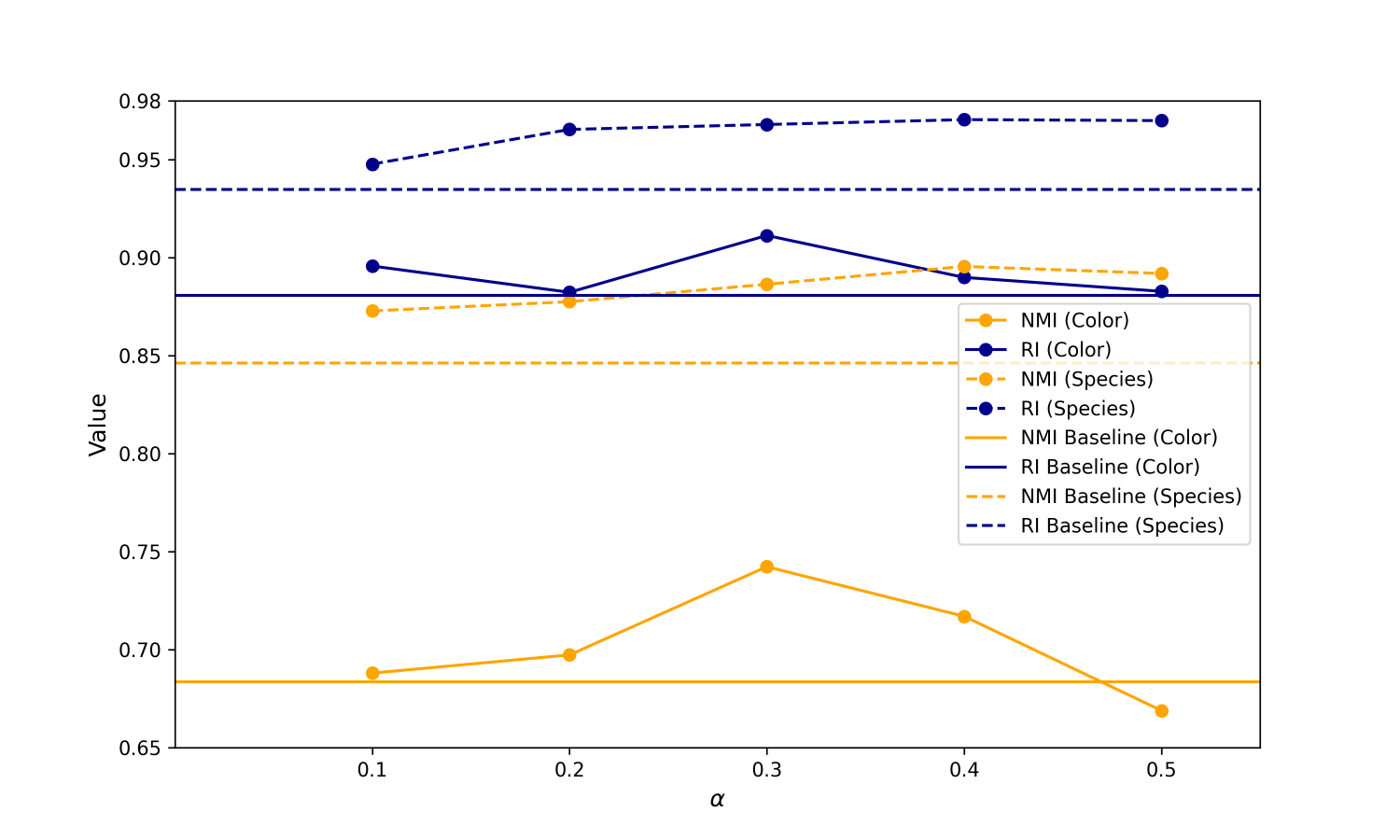}
 \caption{Hyperparameter sensitivity analysis $\alpha$ on Flower dataset}
 \label{hyper}
\end{figure}

Here we conduct the Hyperparameter sensitivity analysis of pseudo label ratio in the Flower dataset with color and species features. As shown in Figure~\ref{hyper}, the NMI and RI scores are generally higher than the baselines(K-means clustering without clustering heads). When $\alpha$ is too high or too low, the performance degrades, mainly due to introducing noisy labels and a lack of supervision, respectively.

\subsection{Target embedding examples}
\label{apd:more_examples}
\begin{figure*}[htb]
    \centering
    \includegraphics[width=\linewidth]{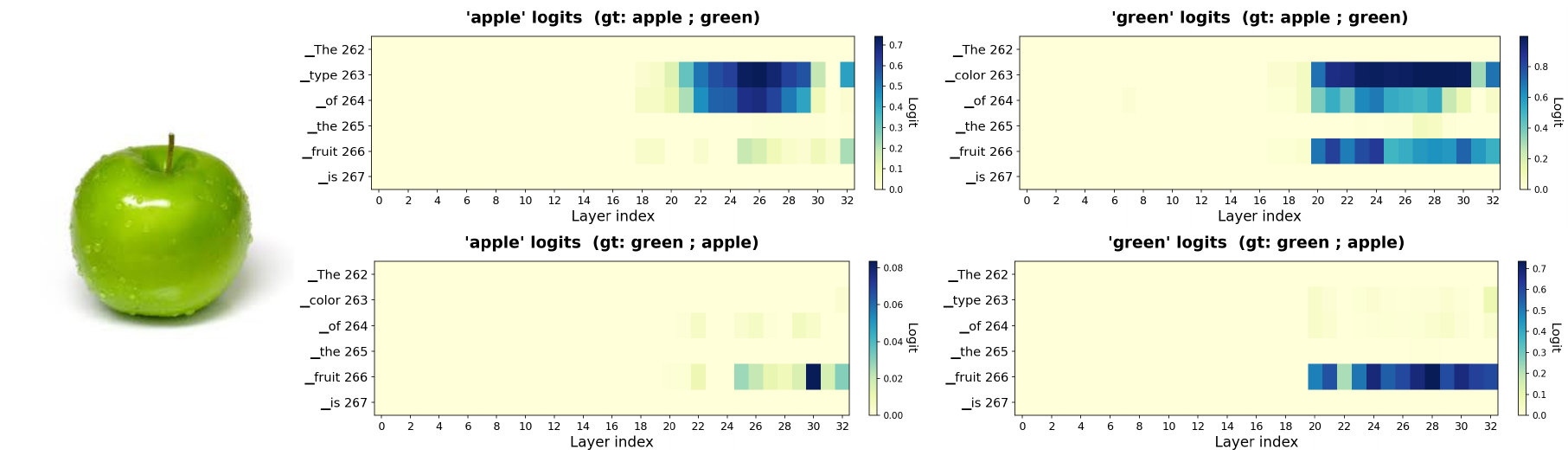}
    \caption{Color and type logits in Fruit dataset for green apple, we choose the embedding $E_{27}(263)$ for both features.}
    \label{figx}
\end{figure*}

\begin{figure*}[htb]
    \centering
    \includegraphics[width=\linewidth]{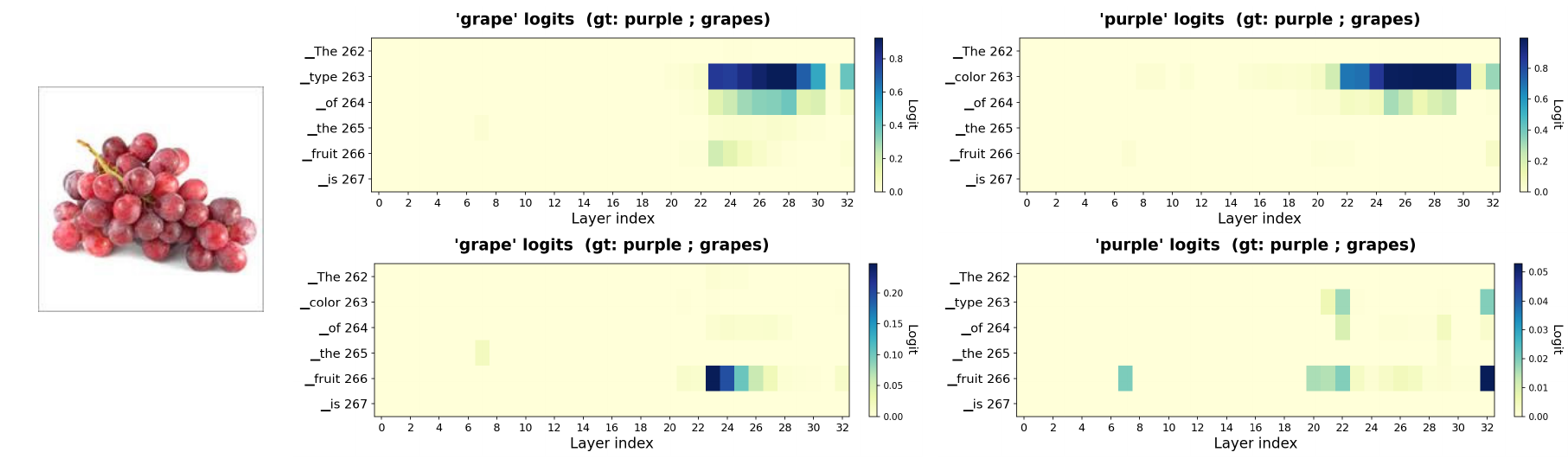}
    \caption{Color and type logits in Fruit dataset for purple grapes, we choose the embedding $E_{27}(263)$ for both features.}
    \label{figxx}
\end{figure*}

Here we provide more examples of target embedding selection with corresponding logits. The high logits are related to feature representative tokens like ``type" and ``color", as shown in Figure~\ref{figx} and Figure~\ref{figxx}. Note that when prompting the model with type features, other visual features, like color features, are not presented at the position of ``type" tokens, facilitating the distinction of different features.

\subsection{Algorithms}
Here we present the pseudo-codes of the algorithms mentioned above.

\begin{algorithm}[H]

\caption{Target Embedding Localization}

\label{alg:1}

\begin{algorithmic}[1]
\STATE \textbf{Input:} prompt\_tokens, number of samples $ N $, threshold $\tau$, selection threshold $\theta$, unembedding\_matrix $W$
\STATE \textbf{Output:} Target embeddings

\FOR{each feature in prompt\_tokens}
    \STATE Sample $ N $ images randomly
    \STATE Generate keywords using GPT-4: \texttt{keywords} $ \leftarrow $ GPT4(``What are common [feature]?")
    \FOR{each keyword in keywords}
        \FOR{each image in sampled\_images}
            \STATE 
            hidden\_states $\leftarrow$\texttt{model.forward(image, text\_prompt)}$\cdot W$ 
            \STATE 
            logit $\leftarrow$ hidden\_states[keyword]
            \FOR{each layer, position in model}
                \IF{logit[layer][position] $ > \tau $}
                    \STATE candidate\_positions[(layer, position)] $\leftarrow$ candidate\_positions[(layer, position)] $ + 1 $
                \ENDIF
            \ENDFOR
        \ENDFOR
    \ENDFOR
\ENDFOR

\STATE max\_count $\leftarrow \max\{$count(p) $\mid$ p $\in$ candidate\_positions\}
\STATE target\_embeddings $\leftarrow $\{embedding(p) $\mid$ p $\in$ candidate\_positions and count(p) = max\_count\}

\STATE \textbf{return} target\_embeddings
\end{algorithmic}
\end{algorithm}

\begin{algorithm}[H]

\caption{Pseudo Label Clustering head}

\label{alg:2}

\begin{algorithmic}[1]
\STATE \textbf{Input:} target\_embeddings, number of clusters $ K $, hyperparameter $\alpha$, maximum training epochs $ \text{max\_epochs} $
\STATE \textbf{Output:} Predictions

\STATE \textbf{Initialize} clusters via K-means on target\_embeddings, pseudo-labeled dataset $ D_{\text{pseudo}} $

\FOR{each cluster $ C $ in clusters}
    \STATE Compute distances $ \leftarrow \{ \|x_i - \text{centroid}(C)\|^2 \mid x_i \in \text{target\_embeddings} \} $
    \STATE Sort samples by distances in ascending order
    \STATE $ D_{\text{cluster}} \leftarrow $Select top $\alpha\%$ samples with smallest distances 
    \STATE $ D_{\text{pseudo}} \leftarrow D_{\text{pseudo}} \cup D_{\text{cluster}} $
\ENDFOR

\STATE \textbf{Initialize} Clustering\_head with input\_dim, hidden\_dim, output\_dim = $ K $
\FOR{epoch = 1 to \text{max\_epochs}}
    \STATE Predictions $ \leftarrow $ Clustering\_head.forward($ D_{\text{pseudo}}.\text{inputs} $)
    \STATE Loss $ \leftarrow $ CrossEntropyLoss(Predictions, $ D_{\text{pseudo}}.\text{labels} $)
    \STATE Update Clustering\_head parameters
\ENDFOR

\STATE \textbf{return} Predictions
\end{algorithmic}

\end{algorithm}

\subsection{Failure Cases Analysis} 
\label{sec:appendix_failure_cases} 
\begin{figure}[htb] 
    \centering
    \includegraphics[width=\linewidth]{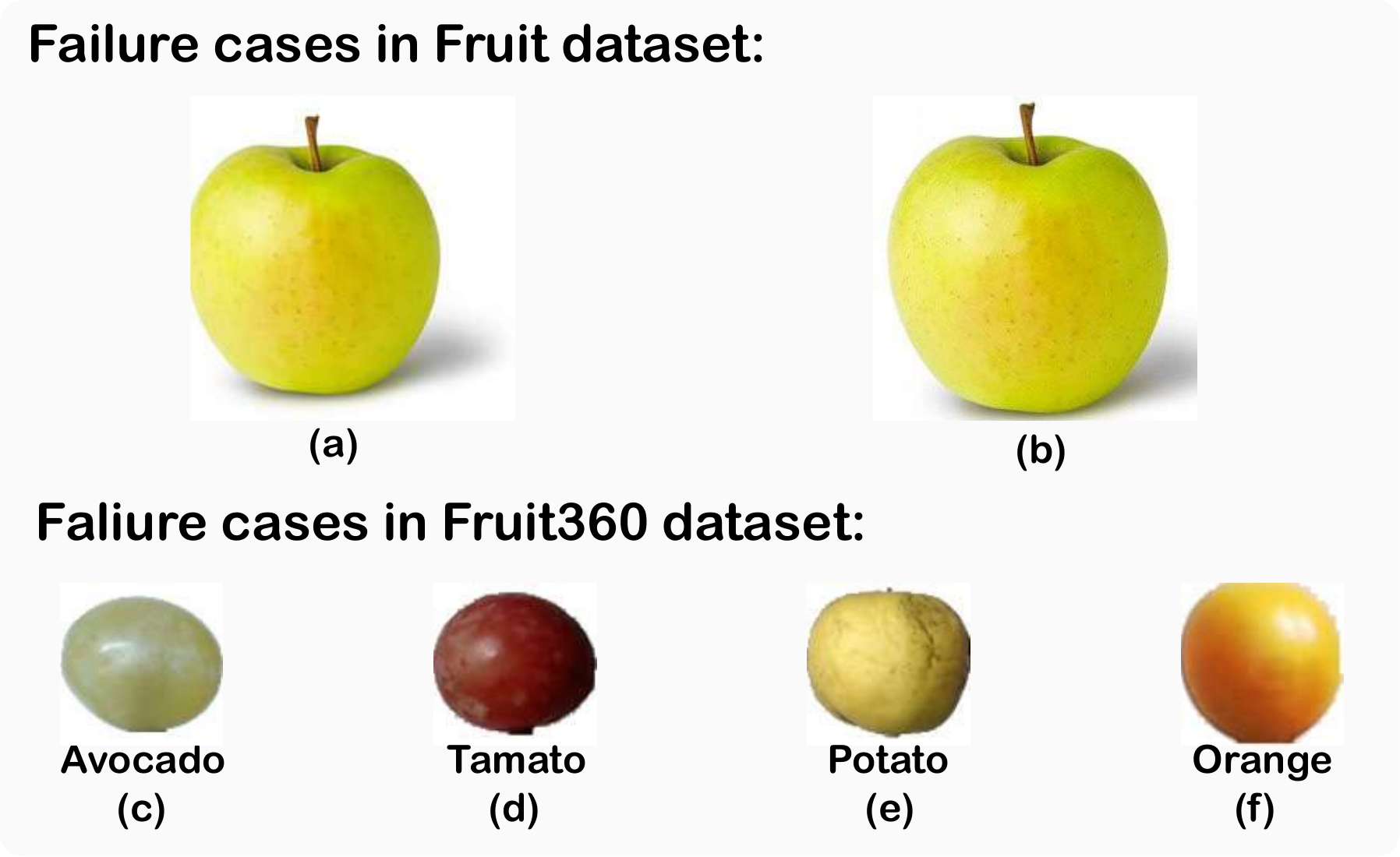}
    \caption{Failure cases in Fruit dataset and Fruit360 dataset}
    \label{fail}
\end{figure}

We elaborate on the minor underperformance observed in the Fruit360-Species NMI and Fruit-Color NMI scores. This issue likely stems from the interplay of unclear image qualities within these datasets and inherent differences in the internal knowledge representations of the models used (i.e., LLaVA and CLIP). As depicted in Figure~\ref{fail}, for instance, a yellow apple (ground truth) is sometimes mistaken as ``Green" (a,b), while other mistakes include identifying a grape as an avocado (c), a grape as a tomato (d), an apple as a potato (e), and a cherry as an orange (f). These specific errors may occur because of the distinct internal knowledge base of LLaVA and CLIP. For example, when directly prompting the models about the color of the aforementioned yellow apple, CLIP responds that it's a yellow apple, whereas LLaVA responds that it's more likely a green apple, a perception which humans might interpret as a yellow-green apple. In contrast, previous methods such as Multi-Map and Multi-Sub exhibit fewer such misidentifications. This is primarily because they are built upon CLIP models and leverage specific keywords (e.g., grapes, apples, oranges, cherries, and bananas) explicitly provided to the model, thereby significantly reducing confusion caused by visually ambiguous images.

\subsection{Prompt Robustness Analysis}
\label{subsec:prompt_robustness}

\begin{table*}
    \centering
    \begin{tabular}{l|c|c|c}
        \toprule
        Prompt Variant & Criterion & NMI & RI \\
        \midrule
        \textit{Original: ``The color of the car is"} & Color & 0.8138 & 0.9235\\
        Variant 1: ``Color of the car:" & Color & 0.8016 & 0.9194 \\
        Variant 2: ``Identify the color of this vehicle." & Color & 0.8143 & 0.9257 \\
        Variant 3: ``Provide the exact car color." & Color & 0.8113 &  0.9268\\
        \midrule
        \textit{Original: ``The brand of the car is"} & Brand & 0.8817 & 0.9589\\
        Variant 1: ``Brand of the car:" & Brand & 0.8784 & 0.9524\\
        Variant 2: ``Tell me the make of the car." & Brand & 0.8911 & 0.9657 \\
        Variant 3: ``What's the brand of the car?" & Brand & 0.8957 & 0.9669 \\
        \bottomrule
    \end{tabular}
    \caption{Prompt Robustness Analysis (NMI and RI on Stanford\_Cars Dataset)}
    \label{tab:pr}
\end{table*}

\begin{figure*}
    \includegraphics[width=\linewidth]{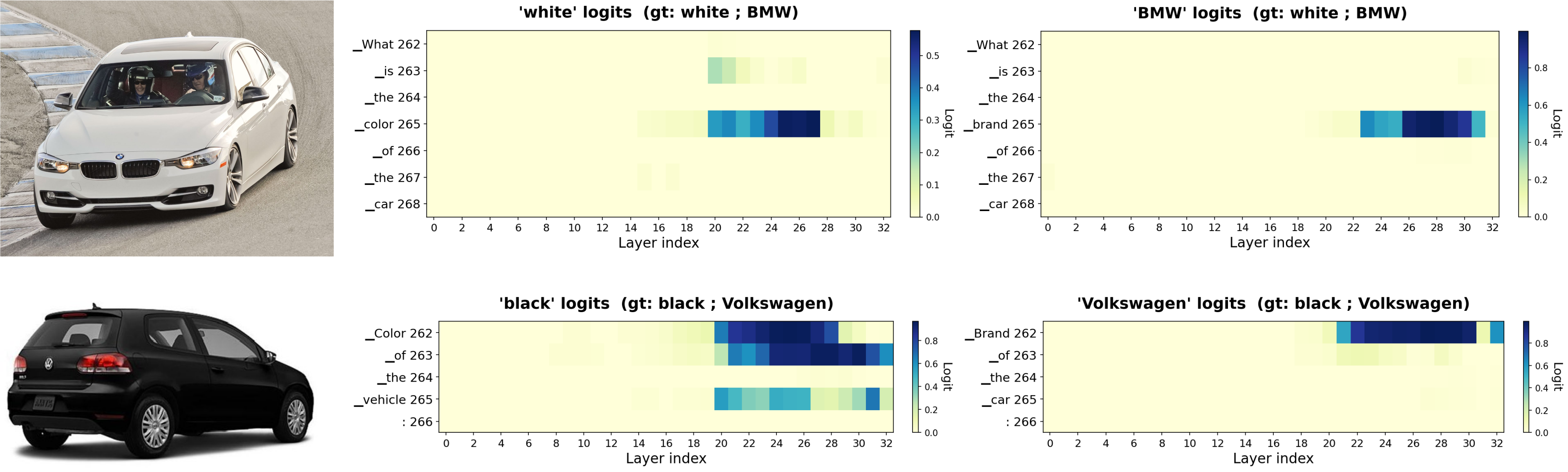}
    \caption{Token logits among different prompts for both color and brand feature.}
    \label{robust1}
\end{figure*}
The performance of methods relying on MLLMs can be sensitive to the precise phrasing of input prompts. To demonstrate the robustness of our proposed ESMC method, particularly concerning the ``Embedding localization" step, this subsection analyzes its performance across various semantically equivalent but lexically diverse prompts. We designed six prompt variants for generating keywords for the ``Color" and ``Brand" criteria. An example set for ``Color" is provided below:
\begin{itemize}
    \item \textbf{Original:} ``The color of the car is"
    \item \textbf{Variant 1:} ``Color of the car:"
    \item \textbf{Variant 2:} ``Identify the color of this vehicle."
    \item \textbf{Variant 3:} ``Provide the exact car color."
\end{itemize}
We applied each prompt variant to the LLaVA model with the criterion of color and brand, ran the full ESMC pipeline, and measured the NMI and RI score. We also present the visualization result in Figure~\ref{robust1}.

As shown in Table~\ref{tab:pr}, the NMI scores exhibit remarkable consistency across all tested prompt variations for the Stanford\_Cars dataset. This strong performance stability confirms the robustness of our ESMC method against minor lexical changes in the prompts used for MLLM's inference. We attribute this robustness to two main factors: firstly, LLaVA's advanced semantic understanding allows it to interpret diverse phrasings with similar intent; and secondly, our ``Embedding Localization" step is designed to identify stable semantic embeddings by leveraging high logits across different keywords with similar meanings.

\subsection{Impact of Sampling Quantity and Keyword Selection}
\label{subsec:sampling}
This section analyzes the influence of sampled image quantity and the choice of Large Language Model (LLM) for keyword generation on the accuracy of the identified target embeddings and the final clustering performance. Note that these modules would only have an impact on the target embedding chosen.

\begin{table}[h!]
    \centering
    \begin{tabular}{c|l|c|c}
        \toprule
        \shortstack{\textbf{Sampled} \\ \textbf{Images}} & 
        \textbf{Criterion} & 
        \shortstack{\textbf{Target} \\ \textbf{Embedding}} & 
        \textbf{NMI}  \\
        \midrule
        3 & Color & $E_{27}(263)$  & 0.8138 \\ 
        3 & Brand & $E_{27}(263)$  & 0.8817 \\
        %\multicolumn{4}{l}{\textit{Special Case: GPT-4 Keywords Miss GT Label}} \\ 
       % 3 & Brand & $E_{18}(263)$ & 0.4531 \\ 
        \midrule 
        5 & Color & $E_{27}(263)$ & 0.8138 \\ 
        5 & Brand & $E_{27}(263)$  & 0.8817 \\
        %\\ 5 & Brand & $E_{21}(263)$ & 0.7833 \\ 
        \midrule
        10 & Color & $E_{27}(263)$  & 0.8138 \\ 
        10 & Brand & $E_{27}(263)$ & 0.8817  \\
        \midrule
        20 & Color & $E_{27}(263)$  & 0.8138 \\ 
        20 & Brand & $E_{27}(263)$ & 0.8817  \\
        \bottomrule
    \end{tabular}
    \caption{Impact of sampled image quantity on target embedding chosen
    and corresponding clustering performance across different criteria.}
    \label{tab:quantity}
\end{table}

\begin{table*}[h!]
    \centering
    \begin{tabular}{l|l|c|c|c} 
        \toprule
        \textbf{LLM} & \textbf{Criterion} & \textbf{LLM Keyword Output} & \textbf{Target Embedding} & \textbf{Final NMI} \\
        \midrule
        \multirow{2}{*}{GPT-4} & Color & \shortstack{Black, White, Silver, Gray, Red, \\ Blue, Green, Gold, Brown, Beige}  & $E_{27}(263)$  & 0.8138  \\
        & Brand & \shortstack{Toyota, Ford, Chevrolet, Honda, Nissan, \\ Volkswagen, Hyundai, Mercedes-Benz, BMW, Audi} & $E_{27}(263)$  & 0.8817  \\
        \midrule
        \multirow{2}{*}{DeepSeek-R1} & Color & \shortstack{White Black Gray Silver Red Blue Green \\ Yellow Orange Brown Gold Beige Bronze Purple Pink}
        & $E_{27}(263)$ & 0.8138  \\
        & Brand & \shortstack{Toyota Honda Ford Chevrolet Volkswagen \\ BMW Mercedes Audi Nissan Hyundai Kia Tesla}  & $E_{27}(263)$ & 0.8817  \\
        \midrule
        \multirow{2}{*}{Gemini-2.5} & Color & \shortstack{White Black Grey Silver Blue Red \\ Green Yellow Brown Beige Purple Gold}  & $E_{27}(263)$ & 0.8138  \\
        & Brand & \shortstack{Toyota Volkswagen Hyundai Kia \\ Ford Honda Nissan Chevrolet BMW Mercedes}  & $E_{27}(263)$ & 0.8817  \\
        \bottomrule
    \end{tabular}
    \caption{Impact of LLM choice on keyword generation, target embedding and clustering performance. The LLMs are prompted with the input: ``What are common car colors(brands)? Provide the answer with single words." The sample size for each feature is 10.}
    \label{tab:llm}
\end{table*}

We conducted experiments on the Stanford\_Cars dataset, evaluating the NMI score as the primary performance metric. We varied the number of sampled images per feature, testing configurations of [3, 5, 10, 20] examples, as illustrated in~\ref{tab:quantity}. We also compared the keyword generation capabilities of three distinct LLMs: GPT-4, DeepSeek-R1 \citep{guo2025deepseek}, and Gemini-2.5 \citep{comanici2025gemini}, as shown in Table~\ref{tab:llm}. For each test, target embeddings were derived using the prompting strategies detailed in Section 3.2.
Our experiments confirm the robustness and high accuracy of our method across varied sampled image quantities and Large Language Model (LLM) choices for keyword generation. Despite this inherent stability, we recommend a sample size of 10-20 images per feature. This suggestion addresses scenarios where generated keywords might not overlap with the ground truth labels, necessitating manual verification of keywords obtained from the LLaVA's output.

\end{document}